\documentclass[letterpaper, 10 pt, conference]{ieeeconf}  

\IEEEoverridecommandlockouts                              

\usepackage{graphicx} 
\usepackage{algorithm}
\usepackage{amsmath}
\usepackage{multirow}
\usepackage{booktabs}
\usepackage{cite}

\title{\LARGE \bf
CHARMS: A Cognitive Hierarchical Agent for Reasoning and Motion Stylization in Autonomous Driving
}

\author{Jingyi Wang, Duanfeng Chu\textsuperscript{*}, Zejian Deng, Liping Lu, Jinxiang Wang and Chen Sun
\thanks{This work is supported in part by the National Natural Science Foundation of China (52472438), the Natural Science Foundation of Hubei Province for Distinguished Young Scholars (2022CFA091), the Key R\&D Program of Hubei Province (2024BAB033), Wuhan Science and Technology Major Project (2022013702025184). ({\it Corresponding author: Duanfeng Chu.})}
\thanks{Jingyi Wang is with the School of Mechanical and Electronic Engineering, Wuhan University of Technology, Wuhan 430063, China
        {\tt\small jxgc\_wangjy@whut.edu.cn}}%
\thanks{Duanfeng Chu is with the Intelligent Transportation Systems Research Center, Wuhan University of Technology, Wuhan 430063, China
        {\tt\small chudf@whut.edu.cn}}%
\thanks{Zejian Deng and Chen Sun are with the Department of Data and Systems Engineering, The University of Hong Kong, Hong Kong SAR 999077, China
        {\tt\small z49deng@hku.hk, c87sun@hku.hk}}%
\thanks{Liping Lu is with the School of Computer Science and Artificial Intelligence, Wuhan University of Technology, Wuhan 430070, China
        {\tt\small luliping@whut.edu.cn}}%
\thanks{Jinxiang Wang is with the School of Mechanical Engineering, Southeast 
University, Nanjing 211189, China.
        {\tt\small wangjx@seu.edu.cn}}
}

\begin{document}

\maketitle
\thispagestyle{empty}
\pagestyle{empty}

\begin{abstract}

To address the challenge of insufficient interactivity and behavioral diversity in autonomous driving decision-making, this paper proposes a Cognitive Hierarchical Agent for Reasoning and Motion Stylization (CHARMS). By leveraging Level-k game theory, CHARMS captures human-like reasoning patterns through a two-stage training pipeline comprising reinforcement learning pretraining and supervised fine-tuning. This enables the resulting models to exhibit diverse and human-like behaviors, enhancing their decision-making capacity and interaction fidelity in complex traffic environments. Building upon this capability, we further develop a scenario generation framework that utilizes the Poisson cognitive hierarchy theory to control the distribution of vehicles with different driving styles through Poisson and binomial sampling. Experimental results demonstrate that CHARMS is capable of both making intelligent driving decisions as an ego vehicle and generating diverse, realistic driving scenarios as environment vehicles. The code for CHARMS is released at https://github.com/chuduanfeng/CHARMS.

\end{abstract}


\section{Introduction}

\begin{figure}[t]
	\centering
	\includegraphics[clip, trim=0.0cm 0cm 0.0cm 0.0cm, width=\linewidth]{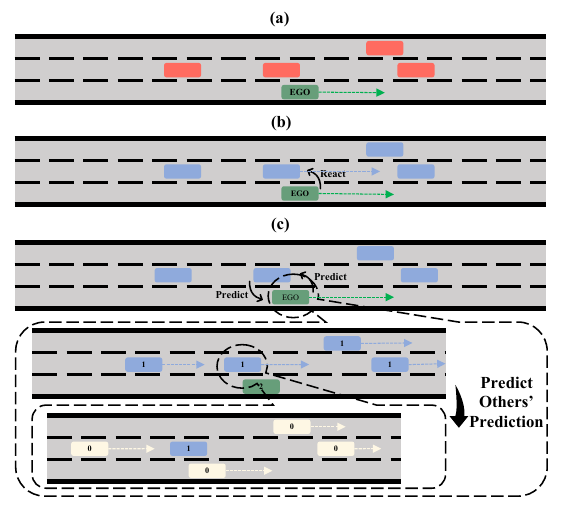}
	\caption{Comparison of Our Approach with Existing Methods. (a) No interaction (rule-based models): The model treats other vehicles as obstacles, and follow a fixed behavior pattern. (b) Unidirectional interaction (game-theoretic or learning-based models): The ego vehicle reacts to the actions of other vehicles, reflecting a one-step reasoning process. (c) Bidirectional interaction (CHARMS): The ego vehicle first predicts how other vehicles anticipate its behavior, then infers their actions based on this prediction, and responds accordingly, exhibiting a two-step reasoning process.}
	\label{fig:comparison}
\end{figure}

Autonomous driving vehicles require sophisticated decision-making models to handle complex interactions with surrounding vehicles in real-world traffic scenarios. Effective modeling of vehicle behavior plays a crucial role in enabling autonomous vehicles to navigate safely and efficiently. Various approaches have been developed to simulate driving behaviors, including rule-based methods and learning-based methods.

Rule-based models, such as the Intelligent Driver Model (IDM)~\cite{treiber2000microscopic} and the MOBIL model~\cite{kesting2007general}, rely on predefined rules to govern vehicle behavior, ensuring predictable and computationally efficient interactions. However, rule-based models tend to yield homogeneous and overly simplistic behaviors, and struggle to represent strategic decision-making and long-term planning, as drivers adjust their behaviors based on a range of cognitive factors and social influences.

Game-theoretic approaches, such as Stackelberg and Nash games, are special kinds of rule-based methods. They offer more interactive models considering the strategic nature of vehicle interactions~\cite{yu2018human, hang2020human, 10810290}. These models introduce more flexibility, allowing agents to make decisions based on the actions of other vehicles. However, they are still limited in terms of cognitive depth, as they typically assume rational behavior at a fixed level. Recent advances have integrated Social Value Orientation (SVO) into game-theoretic models to account for individual preferences. This integration improves variability in driving behaviors, allowing for more diverse and socially nuanced interactions in multi-agent environments~\cite{schwarting2019social, ozkan2021socially, zhang2023game, zhao2024human}. While game-theoretic models improve decision-making dynamics, they are still limited in terms of cognitive depth, as they typically assume rational behavior at a fixed level. 

Learning-based methods have emerged as powerful tools for autonomous driving decision making. Reinforcement learning (RL)~\cite{sallab2017deep, albaba2021driver} enables agents to learn decision policies through trial-and-error interactions with the environment, allowing the discovery of more realistic and dynamic driving strategies. Some reinforcement learning-based methods incorporate behavioral game theory to model the irrational behavior of vehicles~\cite{oyler2016game, li2017game, ma2024evolving}. Imitation learning (IL)~\cite{codevilla2018end} further enhances this by mimicking expert driving behavior, generating human-like actions. However, RL methods often face challenges in aligning rewards with human-like behavior, as learned policies frequently exhibit unnatural or unrealistic actions due to reward shaping biases and the oversimplified nature of simulation environments. In contrast, IL models, while capable of mimicking realistic human driving patterns, typically lack the ability to control behavioral diversity and fail to fully capture the broad spectrum of driving styles observed in real-world scenarios.

Simulators using these models also face limitations. Traffic flow simulators such as SUMO~\cite{krajzewicz2002sumo} and SMARTS~\cite{zhou2010smarts} simplify interactions but fail to capture the full complexity of human decision-making. Ego vehicle-centric simulators like CARLA~\cite{dosovitskiy2017carla} and nuPlan~\cite{caesar2021nuplan} focus on autonomous vehicle performance but often lack diversity and interactivity in background vehicle behaviors.

To address the limitations of these approaches, we propose CHARMS, a decision-making model based on Level-k game theory~\cite{camerer2003cognitive}. The distinction between our approach and the existing methods is illustrated in Fig.~\ref{fig:comparison}. CHARMS incorporates cognitive hierarchy theory to model diverse reasoning depths among agents, coupled with Social Value Orientation (SVO) to capture individual preferences in driving behavior. We employ a two-stage training process consisting of reinforcement learning pretraining and supervised fine-tuning (SFT) to generate decision-making models that exhibit a wide range of human-like driving styles. Additionally, we integrate Poisson cognitive hierarchy (PCH) theory to enable CHARMS to generate more complex simulation scenarios with diverse vehicle styles. The main contributions of this paper can be summarized as follows.

\begin{itemize}
    \item A behavior model integrating Level-k reasoning and SVO is proposed to simulate cognitively diverse driving styles.
    \item A two-stage training scheme (DRL + SFT) ensures both style distinctiveness and behavioral realism.
    \item A scenario generation method based on PCH theory is used to control driving style distributions, with the aim of creating more realistic and behaviorally diverse simulation scenarios.
\end{itemize}

\section{Methodology}

\begin{figure}[t]
	\centering
	\includegraphics[clip, trim=0.0cm 0cm 0.0cm 0.0cm, width=\linewidth]{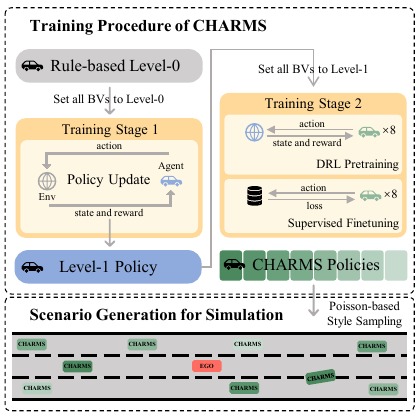}
	\caption{Overall framework of CHARMS. Eight distinct behavior policies are obtained via a two-stage reinforcement learning process followed by supervised fine-tuning, and used to generate diverse and controllable scenarios based on PCH theory.}
	\label{fig:structure}
\end{figure}

This chapter presents the CHARMS model, developed under Level-k game theory and incrementally trained via deep reinforcement learning to achieve Level-2 reasoning: predicting others' prediction. A subsequent fine-tuning stage enhances behavioral realism. The overall framework is shown in Fig.~\ref{fig:structure}.

Specifically, we consider a three-lane highway driving scenario, where vehicles need to reason about the behavior of surrounding vehicles and choose between lane-following or lane-changing actions to achieve a driving state that better aligns with their expectations.

\subsection{Multi-step Reasoning Problem}

Level-k theory does not assume that players always make rational choices, as this assumption is not always valid in single-shot games. It views the game process as a multi-step thinking process: The ego vehicle need to predict the others' driving behavior in order to decide its own, and this process itself involves guessing how the others predict the ego vehicle's behavior.

We assign the vehicle number to be \(i\). The vehicles other than \(i\) are denoted as \(-i\), and the strategy of vehicle \(i\) is represented by \(j\), denoting the strategies of other vehicles. The decision set of vehicle \(i\) is denoted as \(m_i\), and the specific decision of vehicle \(i\) is represented as \(a_i^j\). The utility of vehicle \(i\) is denoted as \(r_i(a_i^j, a_{-i}^{-j})\). The probability that a Level-k vehicle chooses decision \(a_i^j\) is denoted as \(P_k(a_i^j)\). A Level-k vehicle believes that the probability of a level-h vehicle appearing is denoted as \(g_k(h)\). Our objective is to compute the optimal strategy for each level of vehicle, transforming the problem into an MDP problem, and thus obtaining the optimal driving strategy for Level-k vehicles.

Grounded in empirical findings on human overconfidence~\cite{camerer1999overconfidence}, we make a few assumptions: First, vehicles assume that other vehicles have one level lower cognitive ability than themselves, i.e., \( g_k(k-1) = 1 \). This reflects the tendency of individuals to overestimate their own reasoning abilities and to assume that others are neither significantly more nor less competent. Second, each level of vehicle, when making decisions, will select the strategy they believe to be the most optimal, i.e.,

\begin{equation}
P_k(a_i^*) = 1 \iff a_i = \arg \max_{a_i^j} \mathrm{E}_k \left( \mathrm{r}_i(a_i^j) \right)
\end{equation}

Based on the above assumptions, we can express the expected utility of a level-k vehicle using strategy \( a_i^j \) as:

\begin{equation}
E_k \left( r_i(a_i^j) \right) = \sum_{-j=1}^{m_{-i}} r_i(a_i^j, a_{-i}^{-j}) P_{k-1}(a_{-i}^{-j})
\end{equation}

Once the strategy of a level-0 vehicle is predefined, i.e., \( P_0(a_{-i}^{-j}) \) can be calculated, we can compute the optimal strategy iteratively from level-0 to level-\(k\).

\subsection{DRL Pretraining}

We choose the Double DQN network~\cite{van2016deep} to learn the strategies for each level. According to Camerer's experiment, the average number of thinking steps for humans in a game is approximately 1.5 steps~\cite{camerer2003cognitive}. Therefore, we consider Level-2 as the final training level (two-step thinking). During the Double DQN training process, two networks with the same structure but different parameters are used: the main network \(Q(s, a; \theta)\) and the target network \(Q(s, a; \theta^-)\), where \(\theta\) and \(\theta^-\) denote the parameters of the main and target networks. The main network is used to select actions, while the target network is used to calculate the target Q-value. The parameters of the main network \(\theta\) are updated to the target network parameters \(\theta^-\) at fixed step intervals. In the first step, , the ego vehicle learns a Level-1 strategy using \(Q_1(s, a; \theta_1)\), while environment vehicles follow a predefined Level-0 policy. In the second step, the ego vehicle learns a Level-2 strategy using \(Q_2(s, a; \theta_2)\), with Level-1 policies governing the environment vehicles. In both cases, the target Q-value is computed via:

\begin{equation}
y_t = R_t + \gamma Q(s_{t+1}, \arg\max_{a_{t+1}}(s_{t+1}, a_{t+1}; \theta); \theta^-)
\end{equation}

By minimizing the mean squared error loss between the Q-values predicted by the main network and the target value \( y \), the network parameters \( \theta \) are updated through backpropagation.

With the training scheme defined, we now specify the MDP components: the observation space, action space, and reward function.

We construct the observation space \(S\) based on the kinematic features of vehicles in the environment. \(S\) is a \((n, 5)\) matrix representing the states of \(n\) vehicles, with each row containing an existence mask and the vehicle’s position and velocity in both directions.

To simplify the problem, we use a discrete action space with five available decisions: maintain speed, accelerate, decelerate, lane change to the left, and lane change to the right. Actions are mapped to desired speed and lane position, which are then translated into acceleration and steering using proportional control.

The design of the reward function is crucial for training vehicles with different driving styles. In theory, the cognitive abilities of each level of vehicle increase progressively. Therefore, we have designed the following strategies: Level-0 vehicles do not interact with other vehicles and only use a predefined strategy; Level-1 vehicles interact with other vehicles but consider only their own benefits; Level-2 vehicles interact with other vehicles and make decisions based on both their own and the other vehicles' benefits.

The reward function for Level-1 vehicles consists of three components: safety, efficiency, and comfort. The safety reward is expressed as:

\begin{equation}
r_s = 
\begin{cases} 
\begin{aligned}
&0.5 \times \mathrm{clip}_{(0,1)}\left( \frac{TTC_{front}}{3}\right) \\
&+ 0.5 \times \mathrm{clip}_{(0,1)}\left( \frac{TTC_{rear}}{3}\right),
\end{aligned} & L \neq L_t \\
\mathrm{clip}_{(0,1)}\left(\frac{TTC_{front}}{3}\right), & L = L_t
\end{cases}
\end{equation}

Let \( L \) represent the current lane, and \( L_{t} \) represent the target lane. \( {TTC}_{front} \) (time to collision with the front vehicle) and \( {TTC}_{rear} \) represent the time to collision between the vehicle and the front or rear vehicle in the target lane, respectively. The function \( {clip}_{(0,1)} \) restricts values between 0 and 1; any value greater than 1 is set to 1, any value less than 0 is set to 0, and values between 0 and 1 remain unchanged. Since a \( {TTC} \) greater than 3 seconds is considered to indicate a relatively safe driving state, we divide \( {TTC} \) by 3. After applying the clip function, any \( {TTC} \) exceeding 3 seconds results in no change to the safety reward.

The efficiency reward is expressed as:

\begin{equation}
r_e = {clip}_{(0,1)}\left( \frac{v - v_{rmin}}{v_{rmax} - v_{rmin}} \right)
\end{equation}

Where \( v \) represents the current speed of the vehicle, and \([v_{rmin}, v_{rmax}]\) is the speed reward range. The comfort reward \( r_c \) is set based on the consistency between the current action and the previous action. If the actions are consistent, \( r_c \) is 1; otherwise, \( r_c \) is 0. Thus, the total reward function for a Level-1 vehicle is:

\begin{equation}
R_1 = a_1 r_s + b_1 r_e + c_1 r_c
\end{equation}

Since no distinction in driving style is made for Level-1 vehicles, the weights \( a_1 \), \( b_1 \), and \( c_1 \) are assigned fixed values.

In addition to the safety, efficiency, and comfort rewards, Level-2 vehicles also receive a reward based on the benefits to other vehicles. This reward is defined as the increase in the expected acceleration of the rear vehicle and the rear vehicle in the target lane after performing the current action. The desired acceleration \( a_{desire} \) is computed using the IDM model. The reward for the benefits to other vehicles is represented as:

\begin{equation}
r_o = {clip}_{(-3,3)}\left(a_{rear}^{add} \right) + {clip}_{(-3,3)}\left(a_{target\_rear}^{add} \right)
\end{equation}

Where \( a_{rear}^{add} \) and \( a_{target\_rear}^{add} \) represent the increase in the expected acceleration of the rear vehicle in the current lane and the rear vehicle in the target lane, respectively, after the ego vehicle performs the current action. The total reward function for a Level-2 vehicle is expressed as:

\begin{equation}
R_2 = \mathcal{E} (a_2 r_s + b_2 r_e + c_2 r_c) + \mathcal{O} r_o
\end{equation}

Where \( \mathcal{E} \) is the reward weight of the ego vehicle, \( \mathcal{O} \) is the reward weight of the other vehicles, and both are determined by the SVO of the vehicle.

\subsection{Supervised Fine-tuning}

To ensure the behavioral realism of CHARMS, we introduce a supervised fine-tuning stage following reinforcement learning, where the policy network is further trained using human driving trajectories from the HighD dataset~\cite{krajewski2018highd}.

We begin by preprocessing the HighD dataset to extract scenarios that are consistent with the reinforcement learning setting (i.e., three-lane highways) and discard those with insufficient surrounding vehicles. To generate action labels, we classify vehicle behaviors into one-hot vectors aligned with the predefined action space based on longitudinal acceleration and lateral velocity. Since real-world trajectories may involve simultaneous acceleration and lane-changing behaviors, we compute the confidence scores of the two motion components by normalizing longitudinal acceleration and lateral velocity using their respective means. These scores are then used to generate soft labels. The probability formulation is as follows:

\begin{equation}
P_{\text{speed}} = \frac{a_x / \overline{a_x}}{a_x / \overline{a_x} + v_y / \overline{v_y}}
\end{equation}

\begin{equation}
P_{\text{lane}} = \frac{v_y / \overline{v_y}}{a_x / \overline{a_x} + v_y / \overline{v_y}}
\end{equation}

To prevent the model from overfitting to lane-keeping behaviors, we apply undersampling to balance the distribution of action labels. To retain the stylistic characteristics learned during reinforcement learning while adapting to human driving behaviors, we define the fine-tuning loss as the Kullback–Leibler (KL) divergence between the softmax-normalized Q-values and the soft labels, combined with a regularization term that penalizes deviation from the pretrained parameters:

\begin{equation}
\mathcal{L}_{\text{total}} = \text{KL}\left( \text{softmax}(Q(s)) \,\|\, y \right) + \lambda \| \theta - \theta_{\text{pre}} \|_2^2
\end{equation}

\subsection{Scenario Generation with Poisson Cognitive Hierarchy Theory}

The CHARMS models exhibit human-like cognitive abilities and diverse driving styles. To make the scenarios generated based on this model more diverse and realistic, we referenced Camerer's PCH theory to generate scenarios with different styles~\cite{camerer2004cognitive}. According to PCH theory, the cognitive levels of lower-level players follow a Poisson distribution. Consequently, we transplanted the approach by generating scenarios where the number of CHARMS vehicles follows a Poisson distribution along the SVO style dimension and a binomial distribution along the safety-efficiency style dimension. This allows the ratio of different style vehicles to be controlled by two parameters.

The driving style of a CHARMS vehicle is denoted as \( x_n = (x_{1n}, x_{2n}) \in X \), where \( X \) represents the set of all styles. \( x_{1n} \) represents the style along the SVO dimension, and \( x_{2n} \) represents the style along the safety-efficiency dimension. Then, the probability of the Level-2 vehicle's style \( x_n \) is expressed as:

\begin{equation}
P_2(x_n) = \text{poisson}(x_{1n}; \tau) \cdot \text{binomial}(x_{2n}; \beta)
\end{equation}

Based on the two assumptions in Section A, we can express the expected utility of the ego vehicle using strategy \( s_i^j \) in the scenario as:

\begin{equation}
\begin{split}
E_{\text{ego}} \left( r_{\text{ego}} \left( a_{\text{ego}}^j \right) \right) = 
\sum_{j=-1}^{m_{-\text{ego}}} r_{\text{ego}} \left( a_{\text{ego}}^j, a_{-\text{ego}}^{-j} \right) \\
\left\{ 
\sum_{x_n \in X} P_2(x_n) P_{x_n} \left( a_{-\text{ego}}^{-j} \right) 
\right\}
\end{split}
\end{equation}

From the above equation, it is clear that once \( \tau \) and \( \beta \) are determined, the ego vehicle can solve for the optimal strategy, i.e., predict the environment vehicle's prediction and act. Different values of \( \tau \) and \( \beta \) can generate different scenarios, and in these scenarios, the ego vehicle needs to adopt different strategies to achieve the desired reward. This ensures the controllability and diversity of the generated scenarios.

\section{Experiments}

\begin{figure*}[t]
	\centering
	\includegraphics[clip, trim=0cm 0cm 0cm 0cm, width=\linewidth]{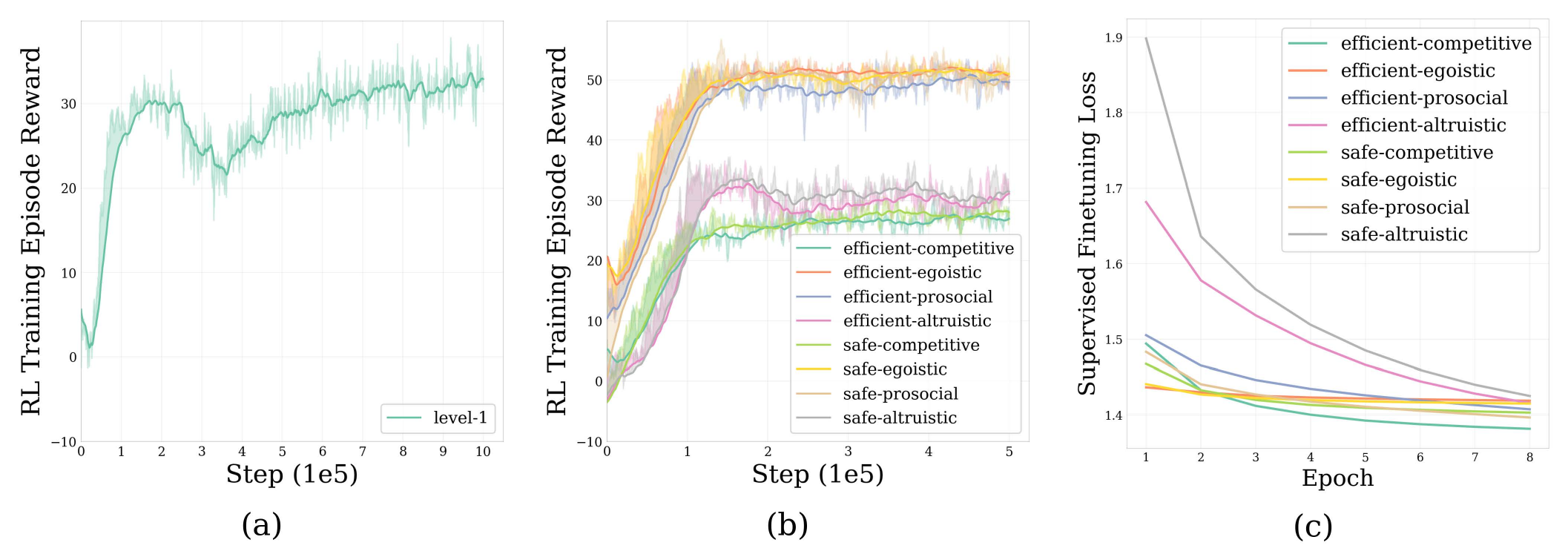}
	\caption{Reward curves of DRL training and loss curves of supervised fine-tuning.}
	\label{fig:training}
\end{figure*}

Experiments were conducted using the highway-env simulator~\cite{Leurent_An_Environment_for_2018}. We begin by presenting the training process of CHARMS, including the hierarchical learning framework, key training parameters, and the convergence behavior observed in the training curves. Following this, we evaluate the final Level-2 strategies from two perspectives. The first is evaluation as the ego agent, which assesses the decision-making capability and the style distinguishability of the trained policies in diverse driving scenarios. The second is evaluation as environment agents, which examines the complexity and realism of the scenarios generated when CHARMS agents are used to simulate surrounding vehicles. We compare CHARMS against several implemented baseline methods, including an IDM+MOBIL model, a Stackelberg game-theoretic model, an imitation learning based model and a prior Level-k model (without SFT on real data).

\subsection{Model Training}

\begin{table}[htbp]
    \caption{Training and Fine-Tuning Hyperparameters}
    \centering
    \renewcommand{\arraystretch}{1.2}
    \begin{tabular}{ll}
    \toprule
    \textbf{Parameter} & \textbf{Value} \\
    \midrule
    RL algorithm               & Double DQN \\
    Policy Network             & MLP $256\times256$ \\
    Learning rate (RL)         & $1 \times 10^{-4}$ \\
    Batch size (RL)            & 32 \\
    Discount factor $\gamma$   & 0.99 \\
    Replay buffer size         & $5 \times 10^4$ \\
    Exploration strategy       & $\epsilon$-greedy \\
    RL training steps (Level-1)& $1 \times 10^6$ \\
    RL training steps (Level-2)& $5 \times 10^5$ \\
    Learning rate (FT)         & $1 \times 10^{-6}$ \\
    Batch size (FT)            & 128 \\
    Epochs (FT)                & 8 \\
    Regularization weight (FT) & 0.2 \\
    \bottomrule
    \end{tabular}
    \label{tab:training_params}
\end{table}

CHARMS is trained through a two-stage process: reinforcement learning (RL) followed by supervised fine-tuning (SFT). The RL phase involves training one Level-1 and eight Level-2 strategy networks using MLPs. Level-2 styles are defined by reward weighting (efficiency vs. safety) and SVO parameters (competitive, egoistic, prosocial, altruistic). In the SFT phase, human driving trajectories from the HighD dataset are used to generate soft action labels, and models are fine-tuned by minimizing the KL divergence between predicted and labeled action distributions. Training hyperparameters are listed in Table ~\ref{tab:training_params}, and learning curves are shown in Fig.~\ref{fig:training}.

The training performance of the RL stage is illustrated in Fig.~\ref{fig:training}(a) and Fig.~\ref{fig:training}(b). The Level-1 model exhibits oscillations during early training (before 400k steps), as the non-interactive behavior of rule-based Level-0 vehicles leads to unstable strategies. In contrast, the Level-2 models trained against Level-1 agents achieve faster and more stable convergence, with most styles stabilizing after 100k steps. To improve robustness and ensure generalization, all Level-2 models were trained for 500k steps.

Fig.~\ref{fig:training}(c) shows the loss curves during the supervised fine-tuning stage. The loss converges rapidly within 8 epochs, owing to the regularization term that constrains deviations from the pretrained parameters. Among all styles, the altruistic agent starts with the highest loss due to its greater behavioral divergence from human trajectories, but still converges to a comparably low value after training.

\subsection{Evaluation as Ego Agents}

The environment was populated by 20 IDM-controlled vehicles with an initial average spacing of 30 m. Five ego-vehicle models: IDM, Stackelberg game-theoretic model, imitation learning-based model, baseline Level-k model, and the proposed CHARMS model with an efficient-egoistic style, were evaluated individually over 1,800 episodes (20s each) for safety, efficiency, and comfort metrics. Results are shown in Table \ref{tab:dmresults}. Safety metrics are omitted for rule-based models due to their inherent safety constraints.

CHARMS substantially outperforms other learning-based models in safety. While slightly less efficient than its learning-based counterparts, it exceeds rule-based models in efficiency. For comfort, CHARMS shows minimal steering variability and significantly fewer harsh acceleration events ($>$2.5 m/s²), occurring at only 7.1\%–29.1\% of the frequency observed in other models. These results underscore its superior driving smoothness and overall decision-making capability as an ego vehicle.

To assess the effectiveness of driving style configurations, eight CHARMS variants were evaluated individually as ego vehicles in the same environment, each over 1,800 episodes (20s each). Behavioral differences are summarized in Table \ref{tab:style_statistics}. Efficiency-oriented models achieved higher average speeds, while safety-oriented models maintained larger average distance headway (DHW) and fewer harsh acceleration events. Competitive and altruistic styles exhibited greater variance in speed and acceleration, reflecting increased responsiveness to surrounding vehicles.

\begin{table*}[t]
    \vspace*{5pt}
    \footnotesize 
    \centering
    \caption{Decision-Making Performance Evaluation of Different EGO Models}
    \begin{tabular}{l c c c c c c}
    \toprule
    \textbf{Ego Model} & \textbf{Collision Rate} & \textbf{TTC$<$3s Frequency} & \textbf{Avg Speed} & \textbf{Acc Std} & \textbf{Yaw Std} & \textbf{Harsh Acc Rate} \\
    \midrule
    IDM+MOBIL   & -     & -     & 23.67 & 1.425 & 0.009 & 0.033 \\
    Stackelberg & -     & -     & 23.73 & \textbf{1.092} & 0.007 & 0.036 \\
    Imitation Learning     & 0.022 & 0.371 & 28.93 & 5.202 & 0.003 & 0.098 \\
    Prior Level-k     & 0.023 & 0.487 & \textbf{32.19} & 5.399 & 0.007 & 0.024 \\
    CHARMS (Ours)      & \textbf{0.004} & \textbf{0.068} & 29.38 & 2.364 & \textbf{0.002} & \textbf{0.007} \\
    \bottomrule
    \end{tabular}
    \label{tab:dmresults}
\end{table*}

\begin{table*}[t]
    \footnotesize 
    \centering
    \caption{Style Distinguishability Evaluation of CHARMS}
    \renewcommand{\arraystretch}{1.2}
    \begin{tabular}{l c c c c c c c}
    \toprule
    \textbf{Ego Model} & \textbf{Avg Speed} & \textbf{Speed Std} & \textbf{Avg Acc} & \textbf{Acc Std} & \textbf{Avg DHW} & \textbf{DHW Std} & \textbf{Harsh Acc Rate} \\
    \midrule
    Efficient-Competitive       & 30.894 & 3.084 & 0.731 & 2.051 & 30.541 & 11.715 & 0.205 \\
    Efficient-Egoistic          & 28.892 & 2.189 & 1.035 & 2.692 & 29.533 & 12.633 & 0.186 \\
    Efficient-Prosocial         & 29.830 & 2.922 & 0.736 & 2.159 & 30.383 & 11.686 & 0.157 \\
    Efficient-Altruistic        & 28.654 & 3.060 & 0.712 & 2.760 & 29.799 & 12.346 & 0.101 \\
    Safe-Competitive            & 29.418 & 3.764 & 0.690 & 2.499 & 31.870 & 12.080 & 0.199 \\
    Safe-Egoistic               & 28.461 & 1.660 & 0.469 & 2.375 & 31.632 & 12.247 & 0.061 \\
    Safe-Prosocial              & 29.703 & 2.708 & 0.321 & 2.336 & 31.406 & 12.560 & 0.061 \\
    Safe-Altruistic             & 27.400 & 3.280 & 0.542 & 2.912 & 30.723 & 12.379 & 0.107 \\
    \bottomrule
    \end{tabular}
    \label{tab:style_statistics}
\end{table*}

\subsection{Evaluation as Environment Agents}

\begin{figure}[t]
	\centering
	\includegraphics[clip, trim=0cm 0cm 0cm 0cm, width=\columnwidth]{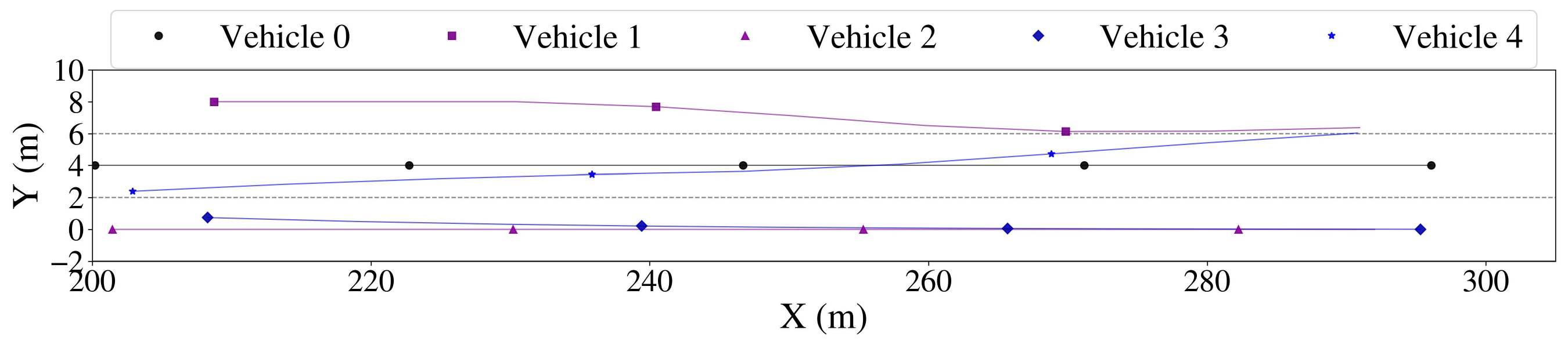}
	\caption{A typical edge case generated by CHARMS with PCH theory.}
	\label{fig:edge_case}
\end{figure}

To evaluate the complexity of scenarios generated by CHARMS as environment agents, we conducted experiments using IDM as the ego vehicle interacting with five types of environments: IDM, Stackelberg, imitation learning, baseline Level-k, and CHARMS with PCH theory. Each setting was run for 1,800 episodes (20s each), and key ego metrics were recorded (Table \ref{tab:complexity}). Interaction density, defined as the average number of interacting vehicles per time step within 30 meters on the same or adjacent lanes, directly reflects the interaction intensity faced by the ego vehicle. Other metrics evaluate the ego vehicle's performance under different environments, and poorer performance implies higher scenario complexity. CHARMS-generated environment induces more frequent behavioral changes in the ego vehicle and includes edge cases with time-to-collision (TTC) less than 3 seconds. An illustrative edge case generated by CHARMS with PCH theory is shown in Fig.~\ref{fig:edge_case}. In this scenario, Vehicle 3 completes an overtaking maneuver on Vehicle 2, while Vehicle 4, after merging into the middle lane, attempts to further change to the leftmost lane. This induces Vehicle 1, which was in the process of merging into the middle lane, to abandon its lane-change maneuver due to the anticipated conflict.

\begin{table*}[t]
    \vspace*{5pt}
    \footnotesize
    \centering
    \caption{Scenario Complexity and Ego Vehicle Stability under Different Environment Models}
    \renewcommand{\arraystretch}{1.2}
    \begin{tabular}{l c c c c}
    \toprule
    \textbf{Env Model} & \textbf{Interaction Density} & \textbf{TTC$<$3s Frequency} & \textbf{Speed Std} & \textbf{Harsh Acc Rate} \\
    \midrule
    IDM+MOBIL         & 2.001 & 0     & 1.177 & 0.076 \\
    Stackelberg       & 2.070 & 0     & 1.185 & 0.086 \\
    Imitation Learning & 1.927 & 0     & 1.814 & \textbf{0.386} \\
    Prior Level-k     & 2.022 & 0     & 1.373 & 0.267 \\
    CHARMS (Ours)            & \textbf{2.167} & \textbf{0.012} & \textbf{2.123} & 0.154 \\
    \bottomrule
    \end{tabular}
    \label{tab:complexity}
\end{table*}

\begin{figure}[t]
	\centering
	\includegraphics[clip, trim=0cm 0cm 0cm 0cm, width=\columnwidth]{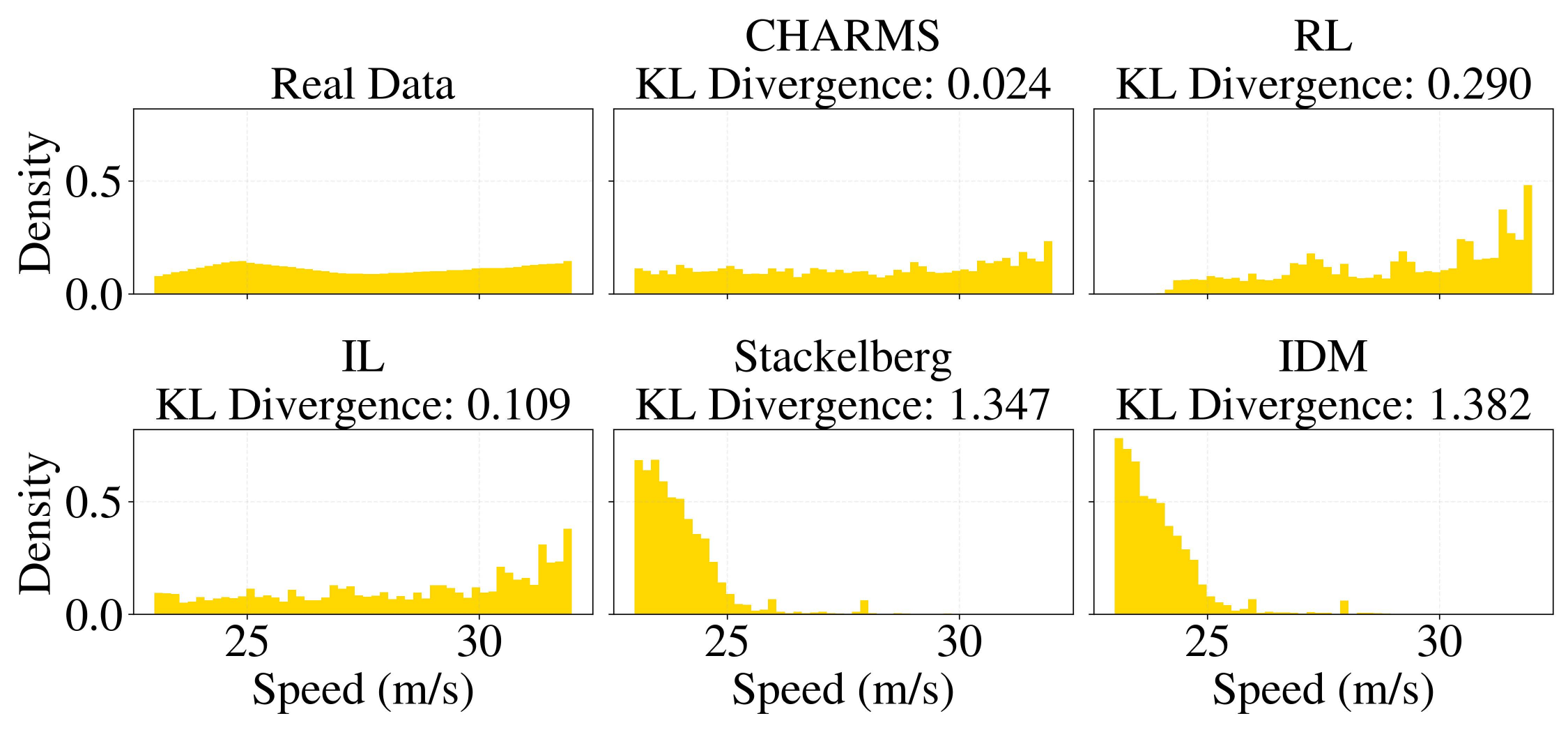}
	\caption{Comparison of speed distributions during car-following across different behavior models.}
	\label{fig:speed_distribution}
\end{figure}

\begin{figure}[t]
	\centering
	\includegraphics[clip, trim=0cm 0cm 0cm 0cm, width=\columnwidth]{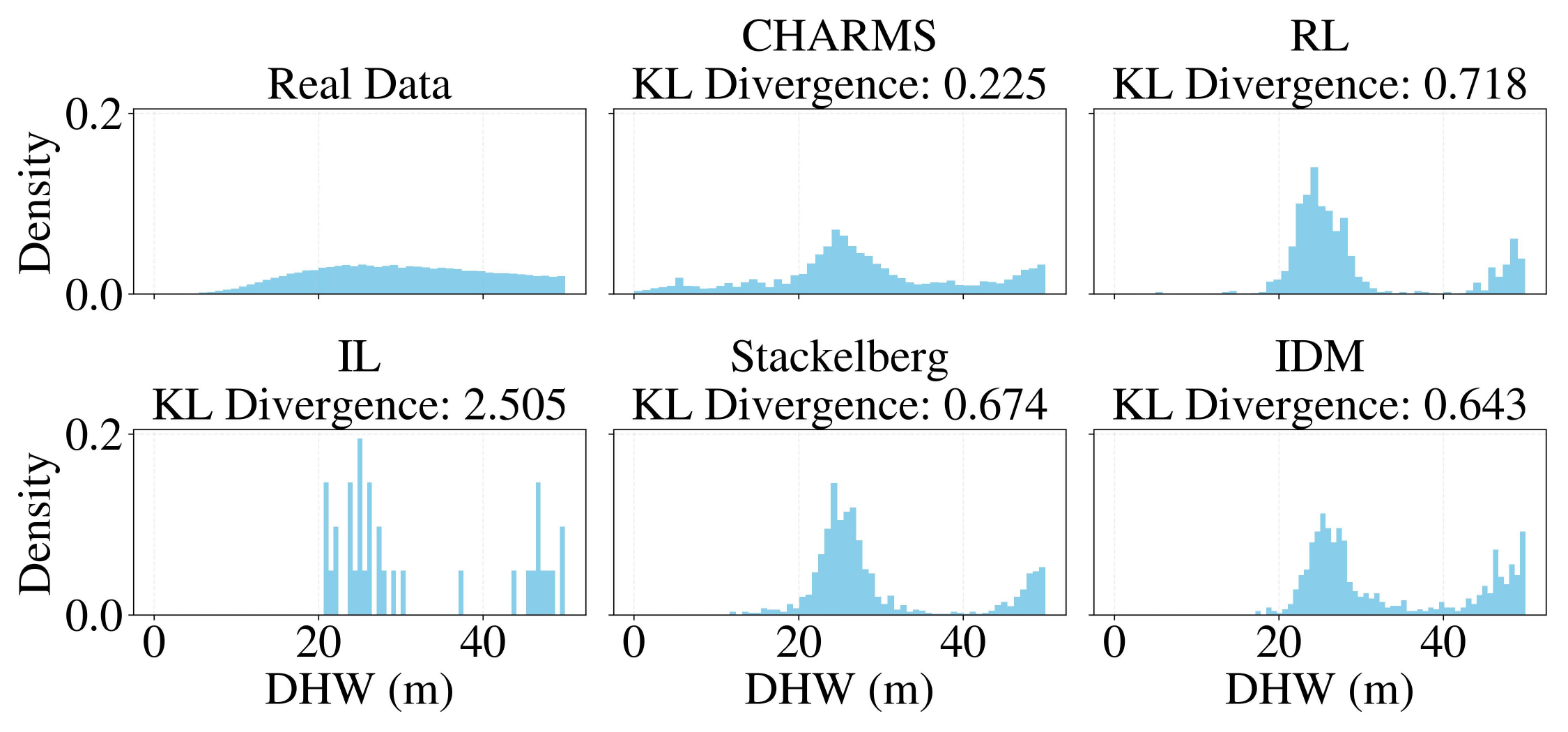}
	\caption{Comparison of DHW distributions during lane changes across different behavior models.}
	\label{fig:dhw_distribution}
\end{figure}

To evaluate the realism of the generated simulation scenarios, we compared scenarios produced by five different models with real human driving data from the HighD dataset. Specifically, we quantified the similarity between each model and real-world data by computing the KL divergence of speed distributions during car-following and DHW distributions during lane changes, as shown in Fig.~\ref{fig:speed_distribution} and Fig.~\ref{fig:dhw_distribution}. The scenarios generated by CHARMS with PCH theory exhibit significantly lower KL divergence from real-world data compared to those generated by other models. This indicates that our approach yields scenarios with more realistic lane-change and car-following behaviors, thereby enhancing the overall plausibility of the simulation environment.

\section{Conclusion}

This paper presents CHARMS, a Level-k-based framework for modeling human-like reasoning in driving decision-making. A set of behavior models, representing diverse human driving styles, is trained through reinforcement learning pretraining followed by supervised fine-tuning. These models exhibit strong reasoning capabilities and demonstrate different social behavior styles. Building upon these behavior models, scenario generation is further structured under the Poisson cognitive hierarchy theory, enabling the systematic assignment of cognitive levels across agents. CHARMS is subsequently integrated into closed-loop autonomous driving simulations, where multi-step reasoning in environment vehicles leads to more diverse behavioral patterns and increased scenario complexity. Experimental results demonstrate that CHARMS outperforms baseline approaches in both decision-making as an ego vehicle and generating complex, realistic scenarios as an environment vehicle model.

Future work will focus on improving the realism of the observation and action spaces and scaling the framework to more complex traffic environments.

\bibliographystyle{IEEEtran}
\bibliography{Literature}

\end{document}